\documentclass[letterpaper, 10 pt, conference]{ieeeconf}
\IEEEoverridecommandlockouts
\usepackage[compress]{cite}
\makeatletter
\let\NAT@parse\undefined
\makeatother
\usepackage[colorlinks=true]{hyperref}
\usepackage{dblfloatfix}
\usepackage[skip=2pt,font=footnotesize]{caption} 
\usepackage{upgreek,bm}
\usepackage{amsmath, amssymb, amsfonts}
\DeclareMathSymbol{\shortminus}{\mathbin}{AMSa}{"39}
\usepackage{graphicx}
\usepackage{adjustbox}
\usepackage{booktabs} 
\usepackage{tabularx}
\usepackage{float} 
\usepackage{color} 
\usepackage[super]{nth}
\usepackage{stmaryrd} 
\allowdisplaybreaks

\usepackage{todonotes}
\usepackage{multirow}
\usepackage{bm}
\usepackage{threeparttable}

\title{\LARGE \bf
Robust 4D Radar-aided Inertial Navigation for Aerial Vehicles
}

\author{Jinwen Zhu$^\dag$, Jun Hu$^\dag$, Xudong Zhao$^\dag$, Xiaoming Lang$^\dag$,  Yinian Mao$^\dag$, and Guoquan Huang$^\dag$$^\ddag$ 
	\thanks{$\dag$ Meituan UAV, Beijing, China (e-mail: {zhujinwen$\mid$ hujun11 $\mid$ langxiaoming $\mid$ zhaoxudong09 $\mid$ maoyinian $\mid$ huangguoquan}@meituan.com).}
 \thanks{ $\ddag$ Dept. of Mechanical Engineering, Computer and Information Sciences, University of Delaware, Newark, DE (email: ghuang@udel.edu).}%
}

\begin{document}

	\newcommand{\tabincell}[2]{\begin{tabular}{@{}#1@{}}#2\end{tabular}}
	
	\maketitle

	\begin{abstract}

While LiDAR and cameras are becoming ubiquitous for unmanned aerial vehicles (UAVs) but can be ineffective in challenging environments, 
4D millimeter-wave (MMW) radars that can provide robust 3D ranging and Doppler velocity measurements are less exploited for aerial navigation.
In this paper, we develop an efficient and robust error-state Kalman filter (ESKF)-based radar-inertial navigation for UAVs.
The key idea of the proposed approach is the point-to-distribution radar scan matching to provide motion constraints with proper uncertainty qualification, which are used to update the navigation states in a tightly coupled manner, along with the Doppler velocity measurements.
%
Moreover, we propose a robust keyframe-based matching scheme against the prior map (if available) to bound the accumulated navigation errors and thus provide a radar-based global localization solution with high accuracy.
Extensive real-world experimental validations have demonstrated that the proposed radar-aided inertial navigation  outperforms state-of-the-art methods in both accuracy and robustness.

\end{abstract}

	\section{Introduction}

As unmanned aerial vehicles (UAVs) are emerging, 
robust and accurate localization and navigation is critical, for example, for safety assurance. 
Camera and LiDAR sensors are among the most widely used for state estimation (e.g.,  VINS~\cite{huang2019visual} and LOAM \cite{zhang2014loam}).
However, these sensors have inherent limitations and may not work well in adverse weather conditions such as rain, fog, and dust storms.
In contrast, millimeter-wave (MMW) radar can perceive such challenging environments and provide robust 3D ranging and Doppler velocity information of detected objects, which in principle is ideal for robust UAV navigation.

While Frequency Modulated Continuous Wave (FMCW) radar has seen significant advancements in recent years (e.g., see \cite{kubelka2023we,yoon2023need,wu2022picking,zhang20234dradarslam}),
it still remains challenging to effectively process its measurements, primarily due to its  sensing  limitations such as large noise in point clouds and sparsity of data.
In particular, in comparison to LiDAR, radar is more susceptible to interference from multipath reflections and other types of noise, leading to larger errors in point cloud measurements. 
Due to its sensing form factor constraints, radar image resolution is low, and the point cloud is extremely sparse (e.g., Continental ARS548  only has about 300 points per frame on average),
which makes accurate frame-to-frame matching almost impossible for high-altitude UAV navigation.

Compared to ground navigation, radar-based SLAM for UAVs entails significantly more challenges.
For example, when UAVs fly at high altitude, the sensor observations are  sparser and the measurement noise is larger. 
The vehicle may often undergo aggressive motions which can easily cause feature matching fail,
and the bird's-eye view can degrade measurements of geometric structure.
In this paper, we will address all these challenges and develop 4D radar-aided inertial navigation to provide efficient, accurate, and robust pose estimation  for UAV navigation. 
In particular, our main contributions are the following:


 
	\begin{itemize}
		\item We develop an efficient and robust ESKF-based  tightly-coupled 4D radar-inertial odometry (RIO) algorithm that fuses Doppler velocity  and scan-to-localmap observations. 
  In particular, we propose a robust point-to-distribution scan matching scheme to enable accurate RIO-based motion tracking.
		
             \item We develop an efficient scheme for matching keyframe scans against a prior map to enable map-based radar localization, which effectively bounds navigation drifts, thus offering high-precision localization solutions.
             
		\item We perform extensive experiments of real-world UAV flights,  showing the proposed approach outperforms existing state-of-the-art algorithms.
		
	\end{itemize}

\begin{figure}[tb]
\centering
\includegraphics[scale=0.28]{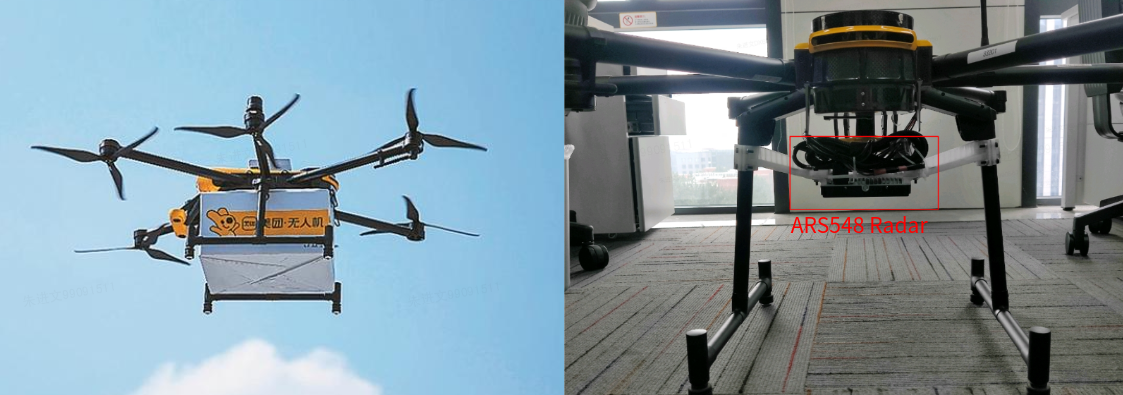}
\caption{Our hardware platform used in urban drone delivery. The radar is mounted underneath the drone, with the FOV facing down. RTK-INS is employed to ensure accurate ground truth for evaluation.}
\label{hardware}
\end{figure}

	\section{Related Work}
 In this section, we provide a brief overview of recent work related to 4D radar(-inertial) odometry. Radar can acquire information about the azimuth, elevation angle, radial Doppler velocity, and distance of the object being measured.
 4D radar odometry algorithms are primarily based on Doppler velocity, point cloud matching, or a combination of them.

Doppler velocity measurements can provide precise radial velocity, and many algorithms utilize this information for state estimation. Doer et al.\cite{DoerMFI2020} use a 3-Point RANSAC Least Squares approach for ego velocity estimation. It requires a single radar scan only making use of the direction and Doppler velocity of each detected object. After this, they proposed EKF-RIO\cite{DoerENC2020}\cite{DoerJGN2022}, based on the EKF framework, fuses IMU with the velocities calculated by radar to determine odometry. Kubelka et al.\cite{kubelka2023we,yoon2023need,wu2022picking,ng2021continuous} also solely utilized Doppler and IMU data for fusion, yielding results that surpassed those achieved with point cloud matching. Hexsel at al. proposed DICP algorithm\cite{hexsel2022dicp}, it derived an observation equation based on Doppler velocity. This equation is solely related to the measured speed of the point and is independent of the spatial structure of the surrounding environment. This helps to address the issue of constraint degradation that ICP faces in unstructured scenarios.
Additionally, the work \cite{nissov2024degradation,kramer2020radar,huang2024multi,hong2021radar} also utilized radar Doppler velocity to address the degradation problem of other sensors.

The RIO system based on Doppler velocity will inevitably produce cumulative errors in position and heading. Many works use point cloud matching to reduce cumulative errors. Zhang et al. proposed 4DRadarSLAM\cite{zhang20234dradarslam}, which employs ADPGICP for scan-to-scan matching to perform trajectory calculations. In the backend, ScanContext\cite{kim2018scan} is used for loop closure detection. Once a loop closure is detected, pose optimization is carried out.
4D iRIOM\cite{zhuang20234d} fusion IMU and radar information based on the filtering framework. It uses the speed information solved by a single frame and IMU for loose coupling fusion, and then updates the point cloud observation in a distribution-to-multidistribution manner. Michalczyk et al.\cite{michalczyk2022tightly} \cite{michalczyk2023multi} use ESKF to tightly couple fuse IMU, Doppler velocity observation and radar point-to-point matching observation. Huang et al.\cite{huang2024less} use IMU as a prior to eliminate dynamic points of radar, use RCS bounded data association for point-to-point matching, and use a factor gragh\cite{loeliger2004introduction} tightly-coupled  fuse  Doppler velocity residuals and scan matching residuals. Xu et al.\cite{xu2024modeling} derived the uncertainties of Doppler and point cloud observations from the sensor's measurement model. They then used these uncertainties for data association and fusion, thereby enhancing the accuracy of RIO.
Besides, there are also some learning-based approaches \cite{lu2020milliego,10160681,yin2021rall} that employ radar for pose estimation.

Doppler velocity constraints can effectively constrain local changes in state, but cannot avoid cumulative drift. Scan matching can reduce cumulative drift, but the noise and sparsity of radar point clouds make matching extremely difficult. In principle, a combination of them can achieve better results.
However, the radar point cloud characteristics  varies significantly across different scenarios. Radar measurements  in the UAV flight scenario are sparser and noisier compared to those on the ground.
Very few radar odometry algorithms are specifically designed for UAV navigation.
To address the issue of sparse and noisy point cloud matching constraints in UAV flight scenarios, we particularly design the RIO system to integrate Doppler observations and scan matching observations in order to achieve better  accuracy.

 \section{Overview of Radar-Inertial Navigation}
 	
  \subsection{System Design}

    \begin{figure}
		\centering
		\includegraphics[scale=0.16]{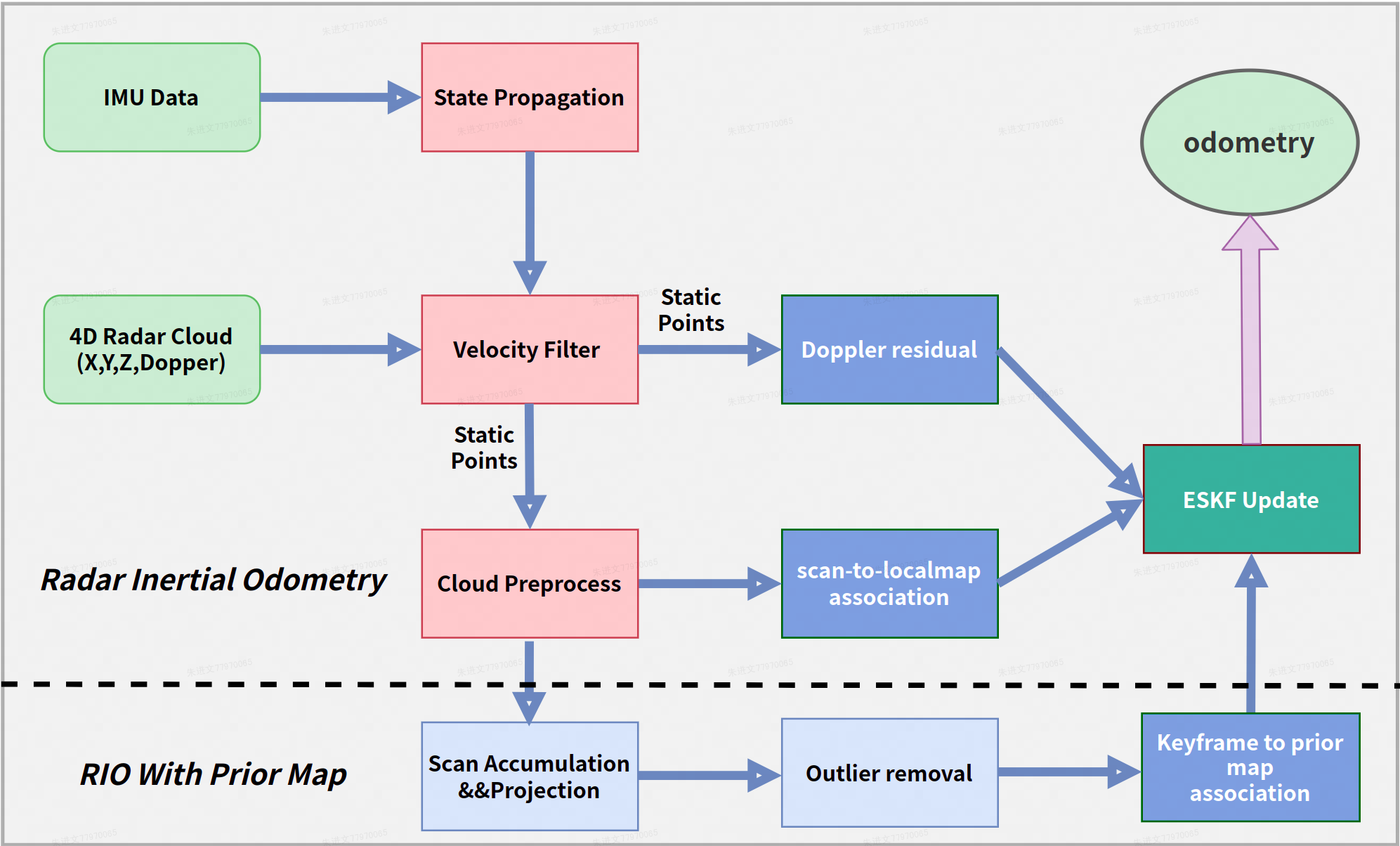}
		\caption{System overview of the proposed radar-inertial navigation system.}
		\label{fig:overview}
	\end{figure}

Inertial navigation serves as the backbone of the proposed radar-inertial navigation system,
which is aided and updated with different radar information including Doppler velocity, scan matching and prior map, within the ESKF framework.
Fig.~\ref{fig:overview} shows the overall architecture of our system.
While the overall system is standard, the proposed robust fusion of 4D radar data is innovative by fully exploiting its sensing characteristics. 

Specifically, 
upon receiving a frame of radar scan, we first verify the Doppler measurements against the state prior and remove outliers, include dynamic and noisy points.
We then calculate the Doppler residual and the point-to-distribution residual between the static points and local-map to update the  state and incrementally update the local map.
If a prior map is available, we match the radar keyframe against the prior map.
We accumulate static point clouds for each frame, and once a while, project the multi-frame point cloud onto the current frame, merging them to create a relatively dense keyframe point cloud. 
Outliers in the keyframe are removed, and point-to-distribution residuals between the keyframe and prior map are built to update the state.
As the prior map established offline is accurate, our radar localization with prior map can effectively eliminate the RIO drifts.


	\subsection{Inertial Navigation as Backbone}
The backbone inertial navigation system (INS) is standard and has the following navigation state:
   \begin{equation} \label{eq:state}
\begin{aligned}
    \mathbf{X} &= [^G\mathbf{R}_I, ^G\mathbf{p}_I, ^G\mathbf{v}_I, \mathbf{b}_g, \mathbf{b}_a, 
    ^I\mathbf{R}_r, ^I\mathbf{t}_r] \\
\end{aligned}
\end{equation} 
where $\mathbf{^GR_I},\mathbf{^Gp_I},\mathbf{^Gv_I}$ respectively represent the attitude, position, and velocity of the IMU body in the world system.
$\mathbf{b}_g, \mathbf{b}_a$ are the biases of the gyroscope and accelerometer, 
and $^I\mathbf{R}_r, ^I\mathbf{t}_r$ are the external parameters from IMU to radar.

Each time when the 6DOF IMU measurement  $\mathbf{u} = [\mathbf{a}_m, \mathbf{\omega}_m]$ is available,
we use it to predict the nominal state and covariance of the error state as follows~\cite{geneva2020openvins}:
\begin{equation}\label{state predict}
\begin{aligned}
            \mathbf{R}_{i+1} &= \mathbf{R}_i \textbf{Exp}(\mathbf{\omega} \Delta{t}) \\
            \mathbf{p}_{i+1} &= \mathbf{p}_i + \mathbf{v}_i \Delta{t} + \frac{1}{2}(\mathbf{R}_i(\mathbf{a}_m - \mathbf{b}_a))\Delta{t}^2 + \frac{1}{2}\mathbf{g}\Delta{t}^2 \\
            \mathbf{v}_{i+1} &= \mathbf{v}_i + \mathbf{R}_i(\mathbf{a}_m - \mathbf{b}_a)\Delta{t} + \mathbf{g}\Delta{t}
\end{aligned}
\end{equation}
The error state covariance is propagated as:
\begin{equation}
{\mathbf{P}}_{i+1}=\mathbf{F}_{{\widetilde{\mathbf{x}}}}{\mathbf{P}}_{i}\mathbf{F}_{{\widetilde{\mathbf{x}}}}^{T}+\mathbf{F}_{{\mathbf{w}}}\mathbf{Q}\mathbf{F}_{{\mathbf{w}}}^{T}
\end{equation}
 where $\mathbf{F}_{\widetilde{\mathbf{x}}}$ and $\mathbf{F}_{\mathbf{w}}$ are the Jacobians with respect to the error state $\mathbf{\widetilde{x}}$ and the noise $\mathbf{w}$ (see \cite{geneva2020openvins}).

If IMU was perfect, the above  inertial propagation should be able to provide accurate navigation solution.
However, in practice, as the MEMS IMUs often used on UAVs are of low quality, 
the backbone inertial navigation requires to be updated or aided by exteroceptive sensors, such as the 4D radar as considered in this work.


\section{Radar-Inertial Odometry}

Building upon the backbone inertial navigation, we will first describe our radar-aided inertial odometry which fuses radar measurements to provide 3D motion tracking. 
 
 \subsection{Doppler Velocity Measurement}
For each radar point, which includes XYZ coordinates and Doppler information, the Doppler velocity of each radar point is treated as an observation. First, we predict the radial velocity of the radar point based on the prior from IMU propagation. Then, we combine it with the radial velocity measured by the radar to obtain the Doppler residual:
 \begin{equation} \label{eq:dopplerResidual}
\begin{aligned}
    \mathbf{Z}_{Doppler}= 
    \underbrace{\frac{(\mathbf{p}_k)^T}{\|\mathbf{p}_k\|}}_{\mathbf{d}(\mathbf{N}_\mathbf{k})\in\mathbb{R}^{1\times3}} \underbrace{\mathbf{^IR_r}^T(\mathbf{R}_i{}^T\mathbf{v}_i+(\hat{\omega}_i-\mathbf{b}_g)^\wedge  {^I \mathbf{t}_r)}}_{\mathbf{K}\in\mathbb{R}^{3\times1}}-v_{dk}
\end{aligned}
\end{equation} 
where $\mathbf{p}_k$ is the 3D position of the radar point, and $v_{dk}$ is the measured Doppler velocity,
and $()^\wedge$ denotes cross product operation.
The position is calculated by the measured azimuth, elevation angle and range, corrupted by white Gaussian noise,
and thus the Doppler residual uncertainty can be derived via covariance propagation.
Note that the Doppler observation only comes from the direction vector of the measurement point and Doppler velocity measurement, 
while the direction vector is only related to the measurement azimuth and elevation angle 
(denoted by $\boldsymbol{\Omega}_k\in\mathbb{S}^2$).
As such, linearizating \eqref{eq:dopplerResidual} with respect to $\boldsymbol{\Omega}_k$ yields:
\begin{equation}\label{eq:doppler_uncertainty}
\begin{aligned}
\mathbf{Z}_{Doppler} \simeq \mathbf{J}_{\mathbf{\Omega}_k}\mathbf{\delta\Omega}_k+\mathbf{n_{doppler}}
\end{aligned}
\end{equation}
The angle noise of radar measurement is represented as $\delta{\boldsymbol{ \Omega_{k}}}$ with covariance 
$\boldsymbol{\Sigma}_{\boldsymbol{\Omega}}=\begin{bmatrix}\sigma_\theta^2&&0\\0&&\sigma_\phi^2\end{bmatrix}$.
As in \cite{xu2024modeling}, by applying perturbations on the manifold, we can derive the noise associated with the direction vector:
\begin{equation}
\mathbf{n}_\mathbf{d}=\underbrace{\begin{bmatrix}\left(\boldsymbol{\Omega}_k\right)^{\wedge}\mathbf{N}(\boldsymbol{\Omega}_k)\end{bmatrix}}_{\mathbf{A}_k}\delta_{\boldsymbol{\Omega}_k}\sim\mathcal{N}\left(\mathbf{0},{}^R\boldsymbol{\Sigma}_k\right)
\end{equation}
where  $\mathbf{N}(\boldsymbol{\Omega}_k) = [\mathbf{N}_1\quad\mathbf{N}_2] \in \mathbb{R}^{3\times2}$ is orthonormal basis of the tangent plane at $\mathbf{\boldsymbol{\Omega}_k}$.
Therefore, the covariance of the Doppler observation is given by:
$  \boldsymbol{\Sigma}_{Z_{doppler}} = \mathbf{J_\Omega} \boldsymbol{\Sigma}_{\boldsymbol{\Omega}}  \mathbf{J_\Omega}^T + \boldsymbol{\Sigma}v_{dk}$.
where $\mathbf{J_{\Omega}} = \mathbf{K}^T\mathbf{A}_k$.

We reject radar points with residuals greater than the $3\sigma$ bound (such as dynamic points and noise points)
and use the Doppler inliers to update the state. 
To this end, the pertinent Jacobians can be computed as follows:
\begin{align}
     \frac{\mathbf{\partial Z_{Doppler}}}{\delta \mathbf{\theta}} &=  -\frac{{\mathbf{p_{k}}}^{T}}{\lVert \mathbf{p_{k}} \rVert}   \mathbf{^IR}_{r}^{T}   ({\mathbf{R_i^T}\mathbf{v_i}})^\wedge \\
     \frac{\partial \mathbf{Z_{Doppler}}}{\delta \mathbf{v}} &=     -\frac{{\mathbf{p_{k}}}^{T}}{\lVert \mathbf{p_{k}} \rVert}   \mathbf{^IR}_{r}^{T}    \mathbf{R_i^T} \\
     \frac{\partial \mathbf{Z_{Doppler}}}{\delta \mathbf{b_g}} &=  -\frac{{\mathbf{p_{k}}}^{T}}{\lVert \mathbf{p_{k}} \rVert}   \mathbf{^IR}_{r}^{T}   \mathbf{^It_r}^\wedge
\end{align}

     \subsection{Scan-to-Localmap  Matching}
Doppler observations do not provide direct constraints on heading and position, which inevitably leads to cumulative errors. Therefore, we employ scan matching constraints to mitigate these cumulative errors.
     
     \subsubsection{Pre-processing Pointcloud} 
     In the process of point cloud matching, we only use the inliers from Doppler update, which effectively avoid dynamic and noisy points.
Next,  we remove points with too low signal-to-noise ratio (SNR) values, as points with low SNR are often unreliable and fail to provide effective constraints. 
The SNR filtering threshold is set based on the statistical results of each point cloud frame, with only the points in the top 95\% of the SNR values kept for scan-to-localmap matching.


 \subsubsection{Point-to-Distribution Matching}
Due to the physical characteristics of the radar, multiple frames of radar scans can hardly observe the same feature point, making it difficult for the point cloud of the current frame to correspond one-to-one with the points of the local map, 
in particular, when experiencing dynamic motions.
Therefore, we do not adopt the point-to-point feature association method but instead use the point-to-distribution feature association method between the current frame and the local map. This approach effectively improves the accuracy and robustness of odometry in high-dynamic UAV flight scenarios.

Specifically, for each static point in the current radar scan, we use the prior to project it onto the map coordinate system, yielding the point $\mathbf{p_w}$, and calculate the covariance of this point based on the noise model. We then use the point $\mathbf{p_w}$ to find the nearest N points in the local map. We consider them independent and fit a Gaussian distribution, yielding the centroid $\mathbf{p_c}$ and covariance $\Sigma \mathbf{p_c}$ for matching association.
Since the points in the local map are not  completely independent, the covariance here needs to be slightly inflated in engineering implementation.
Thus, we can  obtain the residual from point to distribution:
\begin{equation}
     \mathbf{Z_{p_k}} = (\mathbf{R}_i ({^I\mathbf{R}_r}\mathbf{p}_k + ^I\mathbf{t}_r) + \mathbf{p}_i) - \mathbf{p}_c
\end{equation}
The Jacobian of this observation with respect to the state is given by: $\frac{\partial \mathbf{Z_{p_k}}}{\delta \mathbf{\theta}} =  \mathbf{R_i(^IR_rp_k+{^Ip_r})^\wedge} $,$\frac{\partial \mathbf{Z_{p_k}}}{\delta \mathbf{t}} =   - I$.
At the same time, the covariance corresponding to this observation can be calculated as $\mathbf{R_{p_k}} =\mathbf{J_{p_k}}\Sigma\mathbf{p_k} \mathbf{J_{p_k}^T} + \Sigma\mathbf{p_c}  $. Here $\mathbf{J_{p_k}} = \mathbf{R_i} \mathbf{^IR_r}$, and the $\Sigma\mathbf{p_k}$ is the noise covariance of the radar point in the radar frame, as calculated in \cite{xu2024modeling}.
 Based on the residuals and uncertainty of each point cloud observation, we use $\chi^2$ test to reject incorrect matches.

 \subsubsection{Iterated Update}
     As in \cite{xu2021fast}, we use the iterated-ESKF \cite{he2021kalman}  for measurement update. 
     In each iteration,  it is necessary to recalculate the data association of the point cloud and compute the observation residuals and Jacobians to update the state.
     \begin{equation}
    \widehat{\mathbf{x}}_{k}^{\kappa+1} =\widehat{\mathbf{x}}_{k}^{\kappa}\boxplus\left(-\mathbf{K}\mathbf{z}_{k}^{\kappa}-(\mathbf{I}-\mathbf{K}\mathbf{H})(\mathbf{J}^{\kappa})^{-1}\left(\widehat{\mathbf{x}}_{k}^{\kappa}\boxminus \widehat{\mathbf{x}}_{k}\right)\right) 
     \end{equation}
     where $\mathbf{K} =\left(\mathbf{H}^T\mathbf{R}^{-1}\mathbf{H}+\mathbf{P}^{-1}\right)^{-1}\mathbf{H}^T\mathbf{R}^{-1} $.
If the filter converges or the maximum number of iterations is reached, we update the state and covariance as:
\begin{equation}
\bar{\mathbf{x}}_k=\widehat{\mathbf{x}}_k^{\kappa+1}, ~~~\bar{\mathbf{P}}_k=(\mathbf{I}-\mathbf{K}\mathbf{H}) \mathbf{P}
\end{equation}

 \subsubsection{Augmenting Localmap}
     We utilize IKD-Tree\cite{cai2021ikd} to manage the local map point clouds, facilitating efficient expansion and search. Once the scan matching is completed,  the aligned point cloud is inserted into the map, expanding the local map for subsequent point cloud matching.
Moreover, the local map retains only the point clouds in the vicinity of the current location. 
As the vehicle flies, the local map shifts from one to another.

\subsection{System Initialization}
We compute the initial roll and pitch by determining the direction of the average acceleration over a short period. If external position and heading data are available, we use these external sources (such as markers or GPS) to initialize the position and heading. In the absence of external sources, we simply set the initial position and heading to zero.
Finally, we leverage a RANSAC-based least squares method to calculate radar velocity, which is then converted to the world coordinate system to initialize the velocity state.

     \section{Map-based Radar Localization}
Although our RIO presented in the preceding section 
is able to achieve high accuracy, cumulative drift is inevitable. 
Therefore, we introduce a prior map to eliminate cumulative errors and develop a radar-based global localization system.
     
        \subsection{Pointcloud Accumulation and Projection}
The single-frame point cloud generated by the radar is generally very sparse and contains many noise points. Directly matching it with the prior map can easily introduce incorrect constraints. To address the sparsity issue of radar point clouds, we use keyframes accumulated from multiple adjacent frames for point cloud matching.
Since RIO provides accurate local constraints, we use the local relative pose to project the static points of multiple continuous frames onto the latest frame, forming a relatively dense keyframe.
It should be noted that when projecting the point cloud from a past adjacent frame to the current frame, the covariance of the point must also be transformed to the current frame coordinate system for subsequent observation uncertainty calculation.
According to our experiments, dense keyframe matching demonstrates stronger anti-interference capability and higher accuracy compared to single-frame matching.

 \subsection{Outlier Rejection}
Once we obtain a keyframe, given that the radar point cloud typically includes some noise points, we treat outliers as noise points and eliminate them. As the keyframe's point cloud is derived from observations across multiple frames, a point that is markedly distant from other points is likely to be a noise point. Radius filtering is a common method for eliminating outliers, but it typically requires the use of a KD-Tree to search for the number of point clouds within the radius of each point, which is generally time-consuming. Thus, we have designed a voxel-based outlier removal method that can filter out outliers more quickly.
Initially, we project the point cloud into a three-dimensional voxel grid of a fixed size. If there is only one point cloud within a grid, we check whether there are points within the distance threshold of the current point in the point clouds within adjacent voxels. If there are not any, we consider it an outlier and eliminate it. This approach only requires searching for a small number of points and nearby voxels, making it faster.
After removing the outliers, it is beneficial to avoid incorrect point cloud matching constraints.

 \subsection{State Update}
Once we have obtained the keyframes and removed the outliers, we calculate the point-to-distribution distance between the keyframes and the prior map in a manner similar to scan-to-localmap matching. We also compute the Jacobian and the uncertainty and perform an iterated update to correct the current frame's pose in the global map coordinate system. After each keyframe observation update, the keyframes need to be accumulated again. Hence, prior map matching is a low-frequency update and does not affect the real-time performance of the RIO system.

	

	\section{Experimental Results}
In this section, we present our extensive experimental validations, 
including hardware platforms, ground tests, flight tests, ablation study, and real-time performance.

	\subsection{Setup}
 We built the hardware platform based on Meituan's logistics drone, as illustrated in Fig. \ref{hardware}.
 Our drone is equipped with the 4D FMCW Radar ARS548 manufactured by Continental. The radar  is mounted underneath the drone with its Field of View (FOV) facing downward and operates at a frequency band of 77 GHz. It outputs 20 frames of point clouds per second, with each frame containing approximately 200 to 300 points.
 We use the ADS16470 IMU from Texas Instruments in our drone, installed at the centroid  of the drone, and outputs data at 250 Hz.
Additionally, the drone is equipped with an RTK-INS system, using the RTK-INS pose as the ground truth for algorithm evaluation.

To verify the performance of our proposed odometry algorithm, we conducted experiments on our dataset and compared it with several SOTA open-source odometry algorithms. These include the well-implemented EKF-RIO, the SOTA point cloud matching odometry KISS-ICP\cite{10015694}, and the SOTA laser odometry algorithm FASTLIO2\cite{9697912}. We adapted FASTLIO to our radar data for experimental comparison.
We also adapted the open-source 4DRadarSLAM\cite{zhang20234dradarslam} to our data. However, due to the sparsity of our radar data, the algorithm performed very poorly, so we did not include it in our comparisons.

        \subsection{Ground Tests}
 Initially, we handheld the experimental platform to collect ground data, including an indoor office scene and two outdoor park scenes. The indoor scene is a large office building scene. We walked around the inside of the office building with a total length of 142m. One outdoor scene is a roadway within the office park, spanning 207m. The other outdoor scene combines a park road and a plaza, totaling 280m. The collection route for each scene does not overlap, and each starts and ends at the same location, forming a loop. As there's no ground truth, we use the loop closure error for evaluation.
          \begin{figure}[thpb]
		\centering
		\includegraphics[scale=0.35]{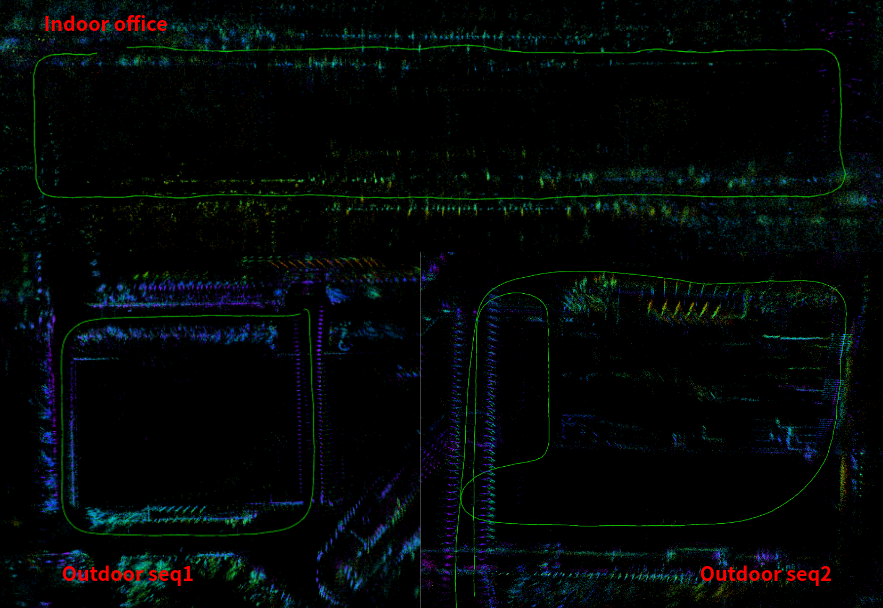}
		\caption{Our RIO in ground data.}
		\label{ground_scene}
	\end{figure}
 	\begin{table}[]
		\caption{Loop closure error(m) of each algorithm on ground radar data}
		\label{loop error on ground data}
		\begin{center}
			\begin{tabular}{c c c c c c}
				\hline
				\tabincell{l}{data} &length & \tabincell{l}{OURS} & \tabincell{l}{EKF-RIO} & KISS-ICP& FASTLIO\\
				\hline
				\tabincell{l}{indoor} &142 & \tabincell{l}{\textbf{4.05}} & \tabincell{l}{11.03} & \tabincell{l}{6.5} & \tabincell{l}{5.66}\\
				\hline
				\tabincell{l}{outdoor seq1} &207& \tabincell{l}{1.88}  & \tabincell{l}{11.11} & \tabincell{l}{2.17} & \tabincell{l}{\textbf{1.28}}\\
				\hline
				\tabincell{l}{outdoor seq2} &280& \tabincell{l}{\textbf{4.87}}  & \tabincell{l}{30.81} & \tabincell{l}{153} & \tabincell{l}{43.8}\\
				\hline
			\end{tabular}
		\end{center}
	\end{table}
 
 The performance of our RIO algorithm on ground radar data is depicted in Fig.\ref{ground_scene}, and the loop closure error is presented in Table \ref{loop error on ground data}.
  From  Table \ref{loop error on ground data}, we can see that  EKF-RIO, KISS-ICP, FASTLIO, and our algorithm can  run on all ground data. Except for outdoor sequence 1, where FAST-LIO performs slightly better than our algorithm, we achieved the highest accuracy.
The indoor and outdoor sequence 1 scenes are well structured, and the point cloud provides good constraints, resulting in similar accuracy for our algorithm, KISS-ICP, and FAST-LIO. In contrast, the geometric structure of the outdoor sequence 2 scene is relatively degraded, causing both KISS-ICP and FAST-LIO to exhibit significant drift. However, our algorithm, which incorporates both Doppler and point cloud constraints, achieves better results. Since EKF-RIO relies solely on Doppler observations and lacks direct observations of heading and position, its error is relatively large across all datasets.

        \subsection{Flight Tests}
\begin{figure*}[tb]
    \centering
     \includegraphics[width=\textwidth]{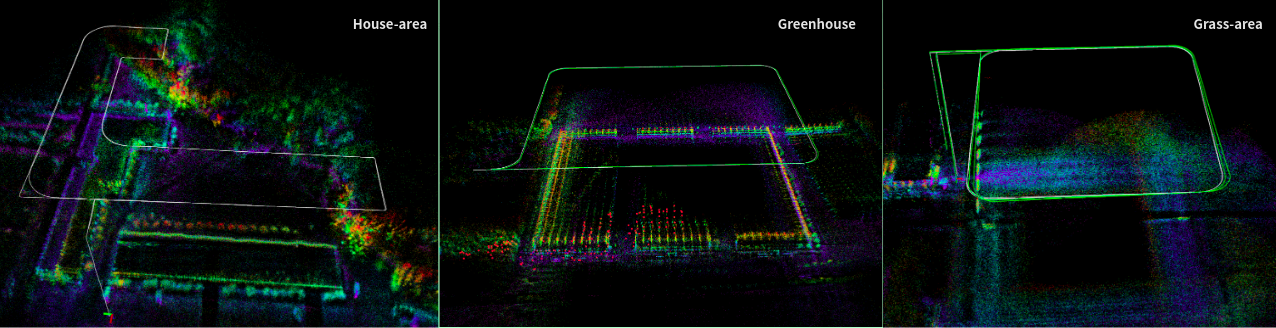} 
    \caption{Our RIO in UAV data.}
    \label{rio_fly}
\end{figure*}

 To verify the performance of our RIO algorithm in UAV flight scenarios, we collected radar data at different scenes and various flight heights. 
The scenes primarily include housing areas, greenhouse areas, and grass areas. The housing and greenhouse areas exhibit relatively strong geometric structures, whereas the grass area lacks structured information, featuring only a large flat ground.
Our flight heights range from 30m to 100m. Due to height restrictions in the greenhouse area, we collected data only at heights below 50m.
We tested EKF-RIO, FAST-LIO, KISS-ICP, and our algorithm on UAV flight data. The performance of our algorithm in the three flight scenes are shown in the Fig. \ref{rio_fly}.
We use RTK-INS as the ground truth to quantitatively evaluate various odometry algorithms, and we use Absolute Pose Error (APE) Root Mean Square Error (RMSE) as the indicator. The unit for APE translation is meters, and the unit for APE rotation is degrees.
Due to the intense movement in drone scenes and the fact that KISS-ICP does not use IMU information, it diverges in all flight data. Additionally, due to the poor structure of the flight scene and the point cloud being overly sparse, FAST-LIO also diverges in some scenes.
As shown in Table \ref{ape uav fly}, our RIO algorithm achieved the highest accuracy in all flight data.
At the same time, our algorithm can stably output odometry in all scenes and at various heights. Notably, even in the grass scene without any geometric structure constraints, the algorithm runs normally without any divergence, demonstrating the best robustness.

         \begin{table}[]
		\caption{APE RMSE of RIO in UAV data}
		\label{ape uav fly}
		\begin{center}
			\begin{tabular}{c c c c c}
				\hline
				\tabincell{l}{Scene/Height } & \tabincell{l}{Length} & \tabincell{l}{OURS} & \tabincell{l}{EKF-RIO} & \tabincell{l}{FASTLIO}\\
				\hline
				\tabincell{c}{House-area\\30m} & \tabincell{l}{870} & \tabincell{l}{\textbf{Tran. 1.137} \\ \textbf{Rot. 1.113}} & \tabincell{l}{Tran. 5.659 \\ Rot. 1.086} & \tabincell{l}{---}\\
				\hline
                \tabincell{c}{House-area\\50m} & \tabincell{l}{910} & \tabincell{l}{\textbf{Tran. 0.984} \\ \textbf{Rot. 0.854}} & \tabincell{l}{Tran. 6.938 \\ Rot. 3.157} & \tabincell{l}{---}\\
				\hline
    \tabincell{c}{House-area\\80m} & \tabincell{l}{970} & \tabincell{l}{\textbf{Tran. 1.553} \\ \textbf{Rot. 1.507}} & \tabincell{l}{Tran. 3.690 \\ Rot. 2.651} & \tabincell{l}{Tran. 3.482 \\ Rot. 4.391}\\
				\hline
    \tabincell{c}{House-area\\100m} & \tabincell{l}{1010} & \tabincell{l}{\textbf{Tran. 0.585} \\ \textbf{Rot. 0.701}} & \tabincell{l}{Tran. 2.945\\ Rot. 3.164} & \tabincell{l}{Tran. 2.834\\ Rot. 3.244}\\
				\hline
    \tabincell{c}{Greenhouse\\30m} & \tabincell{l}{670} & \tabincell{l}{\textbf{Tran. 1.390} \\ \textbf{Rot. 1.389}} & \tabincell{l}{Tran. 5.477 \\ Rot. 2.780} & \tabincell{l}{Tran. 18.371 \\ Rot. 9.852}\\
				\hline
        \tabincell{c}{Greenhouse\\50m} & \tabincell{l}{680} & \tabincell{l}{\textbf{Tran. 0.702} \\ \textbf{Rot. 0.776}} & \tabincell{l}{Tran. 6.334 \\ Rot. 2.247} & \tabincell{l}{Tran. 7.415 \\ Rot. 6.941}\\
				\hline
    
    \tabincell{c}{Grass-area\\30m} & \tabincell{l}{1000} & \tabincell{l}{\textbf{Tran. 3.102} \\ \textbf{Rot. 1.731}} & \tabincell{l}{Tran. 3.379 \\ Rot. 3.224} & \tabincell{l}{---}\\
				\hline
        \tabincell{c}{Grass-area\\50m} & \tabincell{l}{1000} & \tabincell{l}{\textbf{Tran. 2.334} \\ \textbf{Rot. 1.024}} & \tabincell{l}{Tran. 5.587 \\ Rot. 5.372} & \tabincell{l}{---}\\
				\hline
        \tabincell{c}{Grass-area\\80m} & \tabincell{l}{1000} & \tabincell{l}{\textbf{Tran. 2.347} \\ \textbf{Rot. 1.187}} & \tabincell{l}{Tran. 8.355 \\ Rot. 4.701} & \tabincell{l}{---}\\
				\hline
        \tabincell{c}{Grass-area\\100m} & \tabincell{l}{1000} & \tabincell{l}{\textbf{Tran. 3.681} \\ \textbf{Rot. 1.360}} & \tabincell{l}{Tran. 6.384 \\ Rot. 6.425} & \tabincell{l}{---}\\
				\hline

			\end{tabular}
		\end{center}
	\end{table}

	\subsection{Ablation Study}
To compare the impact of Doppler observation, point cloud matching, and different matching methods on RIO, we conducted ablation experiments. The experiments mainly compared the full version of our odometry algorithm with situations where only Doppler observation or only point-to-distribution point cloud matching was used.
As can be seen from Table \ref{ape ablation study}, using only Doppler observation demonstrates good robustness, but due to the lack of direct observation on heading and position, the cumulative error of heading and position is larger.
Using only point cloud matching can provide good constraints in structured scenes, resulting in smaller cumulative error. However, its robustness to different scenes is poor, and it quickly drifts or even diverges in degraded structures like grass scenes.
When combining Doppler observation and point cloud matching, the best accuracy and robustness can be achieved.

Furthermore, we compared the scan matching methods of using only point-to-distribution and point-to-point approaches. As seen from Table \ref{ape ablation study}, the point-to-distribution method of scan matching displays better accuracy and robustness.
This is also related to the radar's characteristics. It is difficult for the radar on the drone to observe the same feature point in multiple scans.

         \begin{table}[]
		\caption{APE RMSE of RIO in Ablation Study}
		\label{ape ablation study}
		\begin{center}
			\begin{tabular}{ccccc}
				\hline
				\tabincell{l}{Scene\\Height } & \tabincell{l}{ours full} & \tabincell{l}{doppler only} & \tabincell{l}{P2D only} & \tabincell{l}{P2P only}\\
				\hline
				\tabincell{c}{House-area\\30m}& \tabincell{l}{\textbf{Tran 1.137}\\ \textbf{Rot 1.113}} & \tabincell{l}{Tran 2.621\\Rot 1.251} & \tabincell{l}{Tran 2.327\\Rot 1.609} & \tabincell{l}{---}\\
				\hline
    				\tabincell{c}{House-area\\50m}& \tabincell{l}{\textbf{Tran 0.984}\\ \textbf{Rot 0.854}} & \tabincell{l}{Tran 2.466\\Rot 1.045} & \tabincell{l}{Tran 1.636\\Rot 0.630} & \tabincell{l}{Tran 3.70\\Rot 0.801}\\
				\hline
    				\tabincell{c}{House-area\\80m}& \tabincell{l}{\textbf{Tran 1.553}\\ \textbf{Rot 1.507}} & \tabincell{l}{Tran 6.329\\Rot 6.839} & \tabincell{l}{Tran 2.923\\Rot 1.068} & \tabincell{l}{Tran 22.87\\Rot 1.401}\\
				\hline
    				\tabincell{c}{House-area\\100m}& \tabincell{l}{\textbf{Tran 0.585}\\ \textbf{Rot 0.701}} & \tabincell{l}{Tran 5.074\\Rot 6.059} & \tabincell{l}{Tran 1.839\\Rot 0.730} & \tabincell{l}{Tran 4.485\\Rot 1.397}\\
				\hline
                    \tabincell{c}{Greenhouse\\30m}& \tabincell{l}{\textbf{Tran 1.390}\\ \textbf{Rot 1.389}} & \tabincell{l}{Tran 5.165\\Rot 3.018} & \tabincell{l}{Tran 2.713\\Rot 1.537} & \tabincell{l}{Tran 5.595\\Rot 2.375}\\
                	\hline
                 \tabincell{c}{Greenhouse\\50m}& \tabincell{l}{\textbf{Tran 0.702}\\ \textbf{Rot 0.776}} & \tabincell{l}{Tran 5.444\\Rot 2.980} & \tabincell{l}{Tran 1.997\\Rot 0.817} & \tabincell{l}{Tran 5.365\\Rot 0.976}\\
                	\hline
                 \tabincell{c}{Grass-area\\30m}& \tabincell{l}{\textbf{Tran 3.102}\\ \textbf{Rot 1.731}} & \tabincell{l}{Tran 4.167\\Rot 2.226} & \tabincell{l}{---} & \tabincell{l}{---}\\
                	\hline
                 \tabincell{c}{Grass-area\\50m}& \tabincell{l}{\textbf{Tran 2.334}\\ \textbf{Rot 1.024}} & \tabincell{l}{Tran 6.764\\Rot 7.021} & \tabincell{l}{---} & \tabincell{l}{---}\\
                	\hline
                 \tabincell{c}{Grass-area\\80m}& \tabincell{l}{\textbf{Tran 2.347}\\ \textbf{Rot 1.187}} & \tabincell{l}{Tran 5.203\\Rot 3.344} & \tabincell{l}{---} & \tabincell{l}{---}\\
                	\hline
                 \tabincell{c}{Grass-area\\100m}& \tabincell{l}{\textbf{Tran 3.681}\\ \textbf{Rot 1.360}} & \tabincell{l}{Tran 5.203\\Rot 3.344} & \tabincell{l}{---} & \tabincell{l}{---}\\
                	\hline

			\end{tabular}
		\end{center}
	\end{table}

\subsection{Map-based Radar Localization}
We utilized RTK-INS as the ground truth to offline construct a radar point cloud map of the UAV flight scenario. This map served as a prior map constraint for testing Radar SLAM in flight scenarios. The experimental results, as shown in Table \ref{radar_global_localization_table}, indicate that our algorithm achieved good positioning accuracy in each tested scenario. In structured scenarios, we were able to achieve decimeter-level positioning. There were no instances of positioning divergence, and the positioning error did not increase over time compared to RIO without a prior map.
 	\begin{table}[]
		\caption{RIO With Prior Map in UAV flight scenario.}
		\label{radar_global_localization_table}
		\begin{center}
			\begin{tabular}{c c | c c}
				\hline
				  \tabincell{l}{Scene/height} & \tabincell{l}{ape rmse} & \tabincell{l}{Scene/height} & \tabincell{l}{ape rmse} \\
				\hline
                \tabincell{l}{House-area 30m} & \tabincell{l}{Tran. 0.388 \\ Rot. 0.576} & \tabincell{l}{Greenhouse 50m} & \tabincell{l}{Tran. 0.416 \\ Rot. 0.638} \\
				\hline
                \tabincell{l}{House-area 50m} & \tabincell{l}{Tran. 0.409 \\ Rot. 0.557} & \tabincell{l}{Grass-area 30m} & \tabincell{l}{Tran. 1.143 \\ Rot. 1.646} \\
            				\hline
                \tabincell{l}{House-area 80m} & \tabincell{l}{Tran. 0.541 \\ Rot. 0.630} & \tabincell{l}{Grass-area 50m} & \tabincell{l}{Tran. 0.661 \\ Rot. 0.653} \\
            				\hline
                \tabincell{l}{House-area 100m} & \tabincell{l}{Tran. 0.446 \\ Rot. 0.576} & \tabincell{l}{Grass-area 80m} & \tabincell{l}{Tran. 2.613 \\ Rot. 1.635} \\
            				\hline
                \tabincell{l}{Greenhouse 30m} & \tabincell{l}{Tran. 0.345 \\ Rot. 0.461} & \tabincell{l}{Grass-area 100m} & \tabincell{l}{Tran. 1.851 \\ Rot. 1.326} \\
            				\hline
			
			\end{tabular}
		\end{center}
	\end{table}
	
\subsection{Real-Time Performance}
To validate the real-time performance of our algorithm, we analyzed the computation time statistics for each module of the algorithm on a consumer-grade PC equipped with an Intel Core i7-11700 2.50GHz CPU and 32GB memory.
As shown in Table \ref{timing_table}, point cloud matching consumes most of the time in the entire algorithmic process. However, our algorithm only requires an average total computation time of 27.82ms per frame, allowing for real-time performance.
	
	\begin{table}[]
		\caption{Timing usage of our RIO}
		\label{timing_table}
		\begin{center}
			\begin{tabular}{l c c c c}
				\hline
				&  \tabincell{l}{ImuPredict} & \tabincell{l}{DopplerFusion} & \tabincell{l}{CloudMatch} & \tabincell{l}{TotalTime} \\
				\hline
				Min(ms)& 0.04 & 0.03 & 0.024 & 0.17\\
				\hline
				Max(ms)& 4.05 & 15.89 & 341.37  & 342.79\\
				\hline
                    Mean(ms)& 0.07 & 0.66 & 26.51 & 27.82\\
				\hline
			\end{tabular}
		\end{center}
	\end{table}

	\section{Conclusions and Future Work}

In this  paper, we have developed  an accurate and robust radar-inertial navigation system for UAVs.
In particular, with the inertial navigation as the backbone, we update the system with the 4D radar Doppler velocity and point-to-distribution pointcloud matching within the efficient ESKF framework.
The proposed approach is also able to effectively utilize the prior map to bound navigation drifts.
Extensive experiments have shown that the proposed method  achieves high accuracy and robustness for UAV navigation. 
Looking ahead, we will integrate the proposed radar-inertial navigation into the autonomous flight system for field tests, while pushing the limit to achieve faster-than-realtime localization solutions.

  \section*{Acknowledgement}
The authors would like to thank Jingyang Wang and Pei Wang for their support related to radar sensors, as well as the colleagues from the Meituan UAV hardware group for their assistance with the drone modification.

	\bibliographystyle{ieeetr}
	\bibliography{mybib}

\begin{thebibliography}{10}

\bibitem{huang2019visual}
G.~Huang, ``Visual-inertial navigation: A concise review,'' in {\em 2019 international conference on robotics and automation (ICRA)}, pp.~9572--9582, IEEE, 2019.

\bibitem{zhang2014loam}
J.~Zhang, S.~Singh, {\em et~al.}, ``Loam: Lidar odometry and mapping in real-time.,'' in {\em Robotics: Science and systems}, vol.~2, pp.~1--9, Berkeley, CA, 2014.

\bibitem{kubelka2023we}
V.~Kubelka, E.~Fritz, and M.~Magnusson, ``Do we need scan-matching in radar odometry?,'' {\em arXiv preprint arXiv:2310.18117}, 2023.

\bibitem{yoon2023need}
D.~J. Yoon, K.~Burnett, J.~Laconte, Y.~Chen, H.~Vhavle, S.~Kammel, J.~Reuther, and T.~D. Barfoot, ``Need for speed: Fast correspondence-free lidar-inertial odometry using doppler velocity,'' in {\em 2023 IEEE/RSJ International Conference on Intelligent Robots and Systems (IROS)}, pp.~5304--5310, IEEE, 2023.

\bibitem{wu2022picking}
Y.~Wu, D.~J. Yoon, K.~Burnett, S.~Kammel, Y.~Chen, H.~Vhavle, and T.~D. Barfoot, ``Picking up speed: Continuous-time lidar-only odometry using doppler velocity measurements,'' {\em IEEE Robotics and Automation Letters}, vol.~8, no.~1, pp.~264--271, 2022.

\bibitem{zhang20234dradarslam}
J.~Zhang, H.~Zhuge, Z.~Wu, G.~Peng, M.~Wen, Y.~Liu, and D.~Wang, ``4dradarslam: A 4d imaging radar slam system for large-scale environments based on pose graph optimization,'' in {\em 2023 IEEE International Conference on Robotics and Automation (ICRA)}, pp.~8333--8340, IEEE, 2023.

\bibitem{DoerMFI2020}
C.~Doer and G.~F. Trommer, ``An ekf based approach to radar inertial odometry,'' in {\em 2020 IEEE International Conference on Multisensor Fusion and Integration for Intelligent Systems (MFI)}, pp.~152--159, 2020.

\bibitem{DoerENC2020}
C.~Doer and G.~F. Trommer, ``Radar inertial odometry with online calibration,'' in {\em 2020 European Navigation Conference (ENC)}, pp.~1--10, 2020.

\bibitem{DoerJGN2022}
C.~Doer and G.~F. Trommer, ``x-rio: Radar inertial odometry with multiple radar sensors and yaw aiding,'' vol.~12, pp.~329--339, 02 2022.

\bibitem{ng2021continuous}
Y.~Z. Ng, B.~Choi, R.~Tan, and L.~Heng, ``Continuous-time radar-inertial odometry for automotive radars,'' in {\em 2021 IEEE/RSJ International Conference on Intelligent Robots and Systems (IROS)}, pp.~323--330, IEEE, 2021.

\bibitem{hexsel2022dicp}
B.~Hexsel, H.~Vhavle, and Y.~Chen, ``Dicp: Doppler iterative closest point algorithm,'' {\em arXiv preprint arXiv:2201.11944}, 2022.

\bibitem{nissov2024degradation}
M.~Nissov, N.~Khedekar, and K.~Alexis, ``Degradation resilient lidar-radar-inertial odometry,'' {\em arXiv preprint arXiv:2403.05332}, 2024.

\bibitem{kramer2020radar}
A.~Kramer, C.~Stahoviak, A.~Santamaria-Navarro, A.-A. Agha-Mohammadi, and C.~Heckman, ``Radar-inertial ego-velocity estimation for visually degraded environments,'' in {\em 2020 IEEE International Conference on Robotics and Automation (ICRA)}, pp.~5739--5746, IEEE, 2020.

\bibitem{huang2024multi}
J.-T. Huang, R.~Xu, A.~Hinduja, and M.~Kaess, ``Multi-radar inertial odometry for 3d state estimation using mmwave imaging radar,'' in {\em 2024 IEEE International Conference on Robotics and Automation (ICRA)}, pp.~12006--12012, IEEE, 2024.

\bibitem{hong2021radar}
Z.~Hong, Y.~Petillot, A.~Wallace, and S.~Wang, ``Radar slam: A robust slam system for all weather conditions,'' {\em arXiv preprint arXiv:2104.05347}, 2021.

\bibitem{kim2018scan}
G.~Kim and A.~Kim, ``Scan context: Egocentric spatial descriptor for place recognition within 3d point cloud map,'' in {\em 2018 IEEE/RSJ International Conference on Intelligent Robots and Systems (IROS)}, pp.~4802--4809, IEEE, 2018.

\bibitem{zhuang20234d}
Y.~Zhuang, B.~Wang, J.~Huai, and M.~Li, ``4d iriom: 4d imaging radar inertial odometry and mapping,'' {\em IEEE Robotics and Automation Letters}, vol.~8, no.~6, pp.~3246--3253, 2023.

\bibitem{michalczyk2022tightly}
J.~Michalczyk, R.~Jung, and S.~Weiss, ``Tightly-coupled ekf-based radar-inertial odometry,'' in {\em 2022 IEEE/RSJ International Conference on Intelligent Robots and Systems (IROS)}, pp.~12336--12343, IEEE, 2022.

\bibitem{michalczyk2023multi}
J.~Michalczyk, R.~Jung, C.~Brommer, and S.~Weiss, ``Multi-state tightly-coupled ekf-based radar-inertial odometry with persistent landmarks,'' in {\em 2023 IEEE International Conference on Robotics and Automation (ICRA)}, pp.~4011--4017, IEEE, 2023.

\bibitem{huang2024less}
Q.~Huang, Y.~Liang, Z.~Qiao, S.~Shen, and H.~Yin, ``Less is more: Physical-enhanced radar-inertial odometry,'' {\em arXiv preprint arXiv:2402.02200}, 2024.

\bibitem{loeliger2004introduction}
H.-A. Loeliger, ``An introduction to factor graphs,'' {\em IEEE Signal Processing Magazine}, vol.~21, no.~1, pp.~28--41, 2004.

\bibitem{xu2024modeling}
Y.~Xu, Q.~Huang, S.~Shen, and H.~Yin, ``Modeling point uncertainty in radar slam,'' {\em arXiv preprint arXiv:2402.16082}, 2024.

\bibitem{lu2020milliego}
C.~X. Lu, M.~R.~U. Saputra, P.~Zhao, Y.~Almalioglu, P.~P. De~Gusmao, C.~Chen, K.~Sun, N.~Trigoni, and A.~Markham, ``milliego: single-chip mmwave radar aided egomotion estimation via deep sensor fusion,'' in {\em Proceedings of the 18th Conference on Embedded Networked Sensor Systems}, pp.~109--122, 2020.

\bibitem{10160681}
A.~Safa, T.~Verbelen, I.~Ocket, A.~Bourdoux, H.~Sahli, F.~Catthoor, and G.~Gielen, ``Fusing event-based camera and radar for slam using spiking neural networks with continual stdp learning,'' in {\em 2023 IEEE International Conference on Robotics and Automation (ICRA)}, pp.~2782--2788, 2023.

\bibitem{yin2021rall}
H.~Yin, R.~Chen, Y.~Wang, and R.~Xiong, ``Rall: end-to-end radar localization on lidar map using differentiable measurement model,'' {\em IEEE Transactions on Intelligent Transportation Systems}, vol.~23, no.~7, pp.~6737--6750, 2021.

\bibitem{geneva2020openvins}
P.~Geneva, K.~Eckenhoff, W.~Lee, Y.~Yang, and G.~Huang, ``Openvins: A research platform for visual-inertial estimation,'' pp.~4666--4672, 2020.

\bibitem{xu2021fast}
W.~Xu and F.~Zhang, ``Fast-lio: A fast, robust lidar-inertial odometry package by tightly-coupled iterated kalman filter,'' {\em IEEE Robotics and Automation Letters}, vol.~6, no.~2, pp.~3317--3324, 2021.

\bibitem{he2021kalman}
D.~He, W.~Xu, and F.~Zhang, ``Kalman filters on differentiable manifolds,'' {\em arXiv preprint arXiv:2102.03804}, 2021.

\bibitem{cai2021ikd}
Y.~Cai, W.~Xu, and F.~Zhang, ``ikd-tree: An incremental kd tree for robotic applications,'' {\em arXiv preprint arXiv:2102.10808}, 2021.

\bibitem{10015694}
I.~Vizzo, T.~Guadagnino, B.~Mersch, L.~Wiesmann, J.~Behley, and C.~Stachniss, ``Kiss-icp: In defense of point-to-point icp – simple, accurate, and robust registration if done the right way,'' {\em IEEE Robotics and Automation Letters}, vol.~8, no.~2, pp.~1029--1036, 2023.

\bibitem{9697912}
W.~Xu, Y.~Cai, D.~He, J.~Lin, and F.~Zhang, ``Fast-lio2: Fast direct lidar-inertial odometry,'' {\em IEEE Transactions on Robotics}, vol.~38, no.~4, pp.~2053--2073, 2022.

\end{thebibliography}

\end{document}